\documentclass[runningheads]{llncs}
\usepackage[T1]{fontenc}
\usepackage{chngcntr}
\counterwithout{figure}{section}
\usepackage{caption}
\usepackage{graphicx}
\usepackage{amsmath}
\usepackage{float}
\usepackage{orcidlink}
\begin{document}

\title{Benchmarking Deep Learning-Based Object Detection Models on Feature Deficient Astrophotography Imagery Dataset}

\titlerunning{Benchmarking Object Detection Models on Astrophotography Dataset}

\author{Shantanusinh Parmar\orcidID{0009-0008-1376-0048}\orcidlink{0009-0008-1376-0048}}
\institute{Department of Information and Communication Technology, Marwadi University, Rajkot, India\\
\email{shantanu.c.parmar@gmail.com}}

\maketitle

\begin{abstract}
Object detection models are typically trained on datasets like ImageNet, COCO, and PASCAL VOC, which focus on everyday objects. However, these lack signal sparsity found in non-commercial domains. MobilTelesco, a smartphone-based astrophotography dataset, addresses this by providing sparse night-sky images. We benchmark several detection models on it, highlighting challenges under feature-deficient conditions.
\end{abstract}

\section{Introduction}
Object detection is a key function within Computer Vision with many models existing to detect different objects in an image. The most common CV model, YOLO~\cite{ref_yolov12}, is an excellent example of versatility and high accuracy of object detection models. With widespread use of these models, many improvements have been made to make detections faster, more accurate, and reduce their compute expense.

\begin{figure}[H]
\centering
\includegraphics[width=0.75\textwidth]{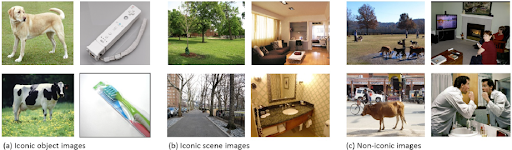}
\caption{Sample of images in the MS COCO dataset~\cite{ref_coco}} \label{COCO_fig}
\end{figure}

\begin{figure}[H]
\centering
\includegraphics[width=0.7\textwidth]{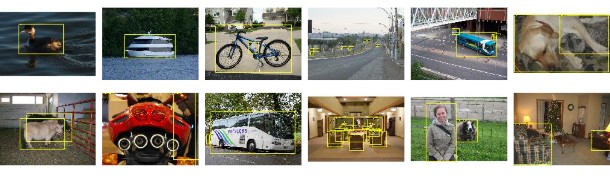}
\caption{Sample of images in the PASCAL VOC 2012 dataset~\cite{ref_pascalvoc}} 
\label{Pascal_fig}
\end{figure}

These improvements, however, have been done based on feature rich datasets such as COCO ~\cite{ref_coco}, ImageNet~\cite{ref_imagenet}, PASCAL VOC~\cite{ref_pascalvoc}, Cityscapes~\cite{ref_cityscapes}, and KITTI~\cite{ref_kitti}.  While these datasets are quite extensive and have a diverse collection of objects for detection, one shared attribute in all of these datasets is that they are detail-rich, featuring bright objects or have a high object-to-background ratio. That is, there is more visible space covered with detectable objects than the background as shown in
Fig.~\ref{COCO_fig}, Fig.~\ref{Pascal_fig}. 

While for many real life applications such as tracking cars in traffic, augmented reality(AR) or facial recognition, these datasets do not replicate the feature sparsity of applications such as astrophotography, microbiology, and satellite imaging. Here, the prominent features are hidden in a swath of background data. By nature, these datasets have very few objects of interest. Consequently, it might be the case that conventional models might not perform well on these datasets.

A notable example demonstrating need for such datasets is the crash of NASA's Ingenuity Mars helicopter on the Martian surface.Leading investigations by NASA suggest that the probable cause of crash was the spacecraft's Vision-aided navigation system misinterpreting the featureless Martian dune terrain as static and hence leading to standy mode activation and ultimately a crash~\cite{ref_NASA}.

Another notable field with high feature-deficiency is Astrophotography. Although used extensively in navigation systems for spacecrafts like JWST (for guide star tracking)~\cite{Open-Star}, telescopes on Earth (especially smaller in size) rely primarily on Sideral tracking. This study might help build future systems for Real-Time celestial object tracking in small-medium telescopes (<2m aperture).\vspace{1em}

\begin{figure}[H]
\centering
\includegraphics[width=0.85\textwidth, keepaspectratio]{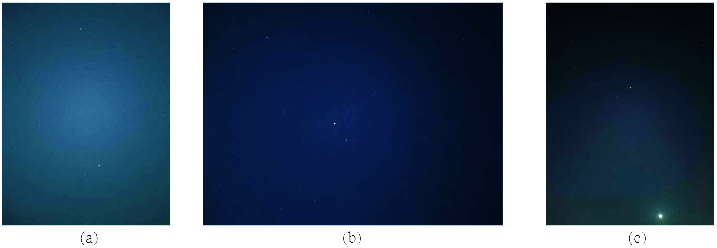}
\caption{Sample images from our dataset, MobilTelesco: a portrait orientation (a),(c) for some images and landscape for some others (b). A prominent source of noise, full moon, in one run of the dataset can be seen in (c).\protect\footnotemark}
\label{MT_fig}
\end{figure}

This paper aims to test and compare 7 common object detection models on a custom astrophotography dataset titled MobilTelesco~\cite{MobilTelesco}. As the name suggests, the dataset is made from capturing astrophotographs through a high-resolution smartphone camera lens (See Fig~\ref{MT_fig} for example image). To the authors' knowledge, this is the first ever smartphone-captured astrophotography dataset. 

\footnotetext{While all images have the same ISO, varying brightness is a change in ambient light (and light pollution) over a period of 4 months.Brightened (78x) for visibility}

The choice of using a smartphone for the images over a conventional astrophotography tool, Digital Single-Lens Reflex (DSLR) cameras, was done to further lower the level of signal to noise ratio as the Active Pixel Sensor in the smartphone is by design not capable of the same performance as a DSLR camera.

\subsection{Related Works}
While data-deficient datasets have been rare in popular vision datasets, some corollary exists. A study on training DnCNN~\cite{ref_nind} Convolutional Neural Networks over Natural Image Noise Dataset (NIND) is a good example. Here, NIND was generated by taking DSLR and smartphone-based images of natural scenarios with varying degrees of ISO setting. The model was trained for blind denoising and performed better than prevalent models such as UNET and BM3D. However, NIND while consisting of higher levels of noise, is not data deficient and consists mainly of natural world object rich in features.

Another notable reference is the 2013 study~\cite{ref_bold} that focuses on using neighboring line segments in features as a bunch and this approach was termed Bunch Of Lines Descriptor (BOLD). This enabled them to identify more features even in sparse feature sets with the caveat of highly curvilinear or simple objects. This shortcoming applies to MobilTelesco directly as seen by section 2.3 in this paper where it is evident most of the features in pixels are sparse, huddled in a few pools across the image and of simple shape.

To address the issue of noisy datasets, there have been many denoising techniques for both images in the dataset as well as labeling and annotations. For noisy images, a low-cost approach for image denoising by Milyaev and Laptev~\cite{ref_Noise_Milyaev} uses edge-preserving filtering, statistics of local binary patterns for the regularity metric, and bilateral filtering where they estimate its intensity smoothing parameters based on the regularity metric. In Li et al.~\cite{ref_Noise_Li}, they two-step denoising method to optimize human-labeled and annotated datasets which tend to have inaccurate labeling and imprecise bounding box coordinates.

Chadwick and Newman~\cite{ref_Noise_Chadwick} investigate the different effects of label noise on the performance of a CNN object detector and showed the use of modified version of co-teaching to mitigate its effects. In our paper, we specifically fill the gap of an in-depth comparison of the most common object detection models on a noisy astrophotography dataset and these denoising techniques could be applied in future studies.

\section{MobilTelesco Dataset}
\subsection{Dataset Creation}
The MobilTelesco dataset~\cite{MobilTelesco} comprises astrophotography images captured using a 50 MP Sony LYT-600 OIS sensor (1/1.95" optical format, 0.8$\mu m$ pixel size) ~\cite{ref_sony_mobile_sensor}. Originally stored in .dng (digital negative) format, the images were converted to .jpg using a lossless pipeline. Each image has a resolution of 3072×4096 pixels with a 24-bit depth.

\subsection{Data Collection and Structure}
A total of 5,400 images were collected over 54 sessions, with each session consisting of two sets of 50 images, spanning 30 minutes per set. Images were captured at intervals of 20s, 10s, and 1s, corresponding to exposures of 10s, 20s, and 30s respectively. These sessions were distributed across five months, covering varying levels of light pollution and seasonal sky brightness (e.g., higher luminosity in summer vs. winter). Image orientations include landscape, portrait, and tilted (<45°) formats.

Each image depicts part of the night sky and includes at least four of the following eight labeled celestial objects: Betelgeuse, Jupiter, Aldebaran, Pleiades cluster, Bellatrix, Zeta Tauri, Elnath, and Hassaleh. Annotations were applied per image and per objext with the dataset being split, 70:15:15, into train, test, and evaluation sets. On average, 3574 labeled instances per class are available, though some frames exclude certain objects due to cropping or occlusion.

\begin{figure}
\centering
\includegraphics[width=0.8\textwidth]{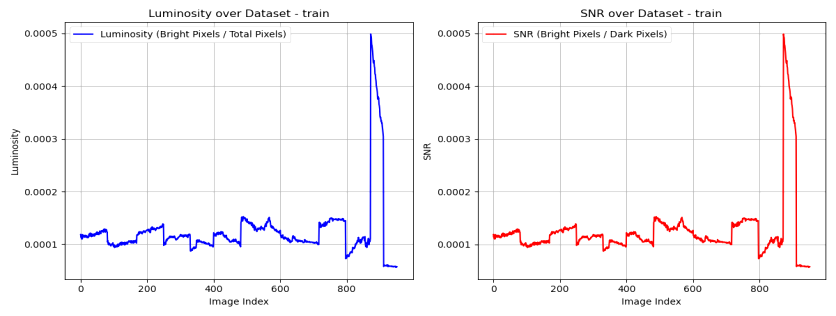}
\caption[Signal-to-Noise Ratio (SNR) and luminosity distribution over the images in MobilTelesco dataset.]{Signal-to-Noise Ratio (SNR) and luminosity distribution over the images in MobilTelesco dataset.\protect\footnotemark}
\label{fig:SNR_lum_fig}
\end{figure}
\footnotetext{The spike indicates the section in dataset where the moon was visible as evident with the high relative diff but low absolute value change in intensity and also from the index.}

\subsection{Dataset Statistics}
The MobilTelesco dataset exhibits high background noise, with celestial bodies occupying only a small fraction of image pixels. The average Signal-to-Noise Ratio (SNR) across the dataset is 0.015\% (Fig.~\ref{fig:SNR_lum_fig}), highlighting the low visibility of target objects.


\begin{table}
\centering
\scriptsize
\caption{Statistical Summary of Pixel Analysis}
\label{pixel_tab}
\begin{tabular}{|c|c|c|c|c|}
\hline
\textbf{Method} & \textbf{Box pixels} & \textbf{Bright pixels} & \textbf{Dark pixels} & \textbf{Average Intensity (\%)} \\
\hline
Mean & 4838 & 26 & 4812 & 25.01587 \\
\hline
Std.Deviation & 5169 & 33 & 5174 & 7.96097 \\
\hline
Variance & 26725757 & 1093 & 26769833 & 63.37704 \\
\hline
Median & 3025 & 17 & 3007 & 23.99935 \\
\hline
1st-Quartile & 2050 & 5 & 2025 & 18.89329 \\
\hline
3rd-Quartile & 5092 & 29 & 5040 & 30.37756 \\
\hline
\end{tabular}
\end{table}

\begin{figure}
\centering
\includegraphics[scale=0.35]{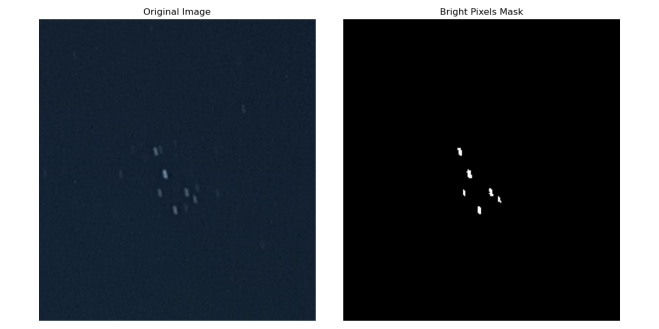}
\caption{Comparison of SNR over selecting factor, Pleiades cluster.}
\label{Pleiades_fig}
\end{figure}

The SNR threshold was tuned to ensure the least bright object, the Pleiades cluster, remains at least 50\% visible after threshold-based masking (Fig.~\ref{Pleiades_fig}). Both localized (bounding box-level) and global SNR scores remain low, underscoring the sparsity of informative features.

To ensure unbiased training, data splits (train/val/test) were randomized using Python's random.shuffle(), based on the Mersenne Twister (MT19937) algorithm~\cite{ref_rng_parallel}. The distribution of light pixels (pixel values above threshold) across labeled objects shows minor variation (see Fig.~6), indicating consistency in object visibility across classes.

\begin{figure}[H]
  \centering
  \begin{minipage}{0.80\textwidth}
    \centering    \includegraphics[width=\linewidth]{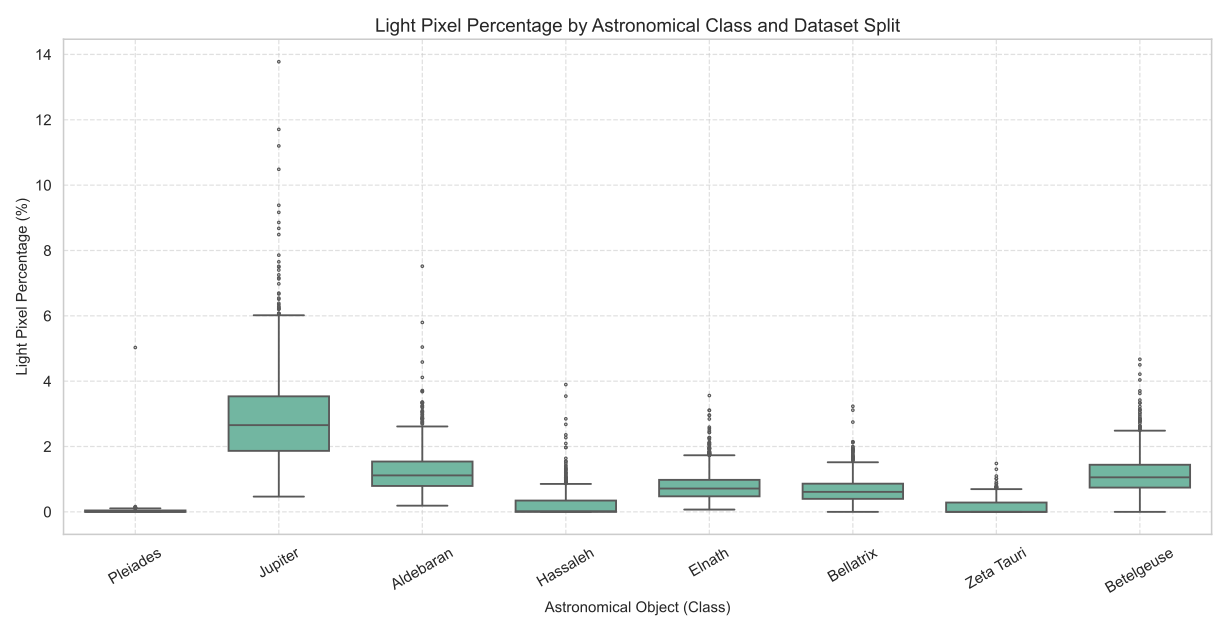}
  \end{minipage}
  \caption{Luminosity distribution per class of Mobiltelesco dataset.}
  \label{fig:light_fig}
\end{figure}

\section{Experimental results \& Analysis}
The models evaluated in this study were selected based on three criteria:
(i) architectural complexity (e.g., number of detection stages),
(ii) underlying frameworks (e.g., Detectron2, PaddlePaddle), and
(iii) reported mean Average Precision (mAP) on the COCO benchmark (see Table~\ref{tab:performance}).
MobilTelesco-specific configurations for each model are detailed in Table~\ref{tab_architecture}. For broader benchmarking comparisons, refer to Table~\ref{tab_mAP}.

Epoch selection was guided by convergence behavior in training loss (see addendum for loss plots). Batch sizes were limited by available computational resources; values beyond 16 images per batch either yielded diminishing mAP gains or caused out-of-memory failures.
\subsection {Training hardware}
All models except YOLOv12 were trained on an Intel Core i7-13700F CPU (16 cores, 24 threads, 2.1 GHz) and NVIDIA RTX 4060 GPU (8 GB, 115W TDP).
YOLOv12 was trained separately on an AMD Ryzen 5 5500 CPU (6 cores, 12 threads, 3.6 GHz) and NVIDIA RTX 3050 GPU (8 GB, 130W TDP).

A complete summary of each model’s performance on the MobilTelesco dataset is presented in Table~\ref{tab:performance}.


\begin{table}[h]
\centering
\caption{Frameworks, architecture stages, and training epochs of models used.}
\label{tab_architecture}
\begin{tabular}{|c|c|c|c|}
\hline
\textbf{Model} & \textbf{Framework} & \textbf{Architecture Stage(s)} & \textbf{Epochs} \\
\hline
Single Shot Detector (SSD300) & PyTorch & Single & 100 \\
\hline
RetinaNet & Detectron & Single & 300 \\
\hline
Faster R-CNN & Detectron & Double & 300 \\
\hline
YOLOv12 & PyTorch & Double & 100 \\
\hline
PP-YOLOE+ & PaddlePaddle & Light & 100 \\
\hline
NanoDet & Independent & Lightweight & 300 \\
\hline
Sparse R-CNN & Detectron & Transformer & 6000 \\
\hline
\end{tabular}
\end{table}

\subsection{Single Shot Detector SSD300}

The SSD300 ~\cite{ref_ssd}model was trained for 100 epochs with a batch size of 32 using a ResNet-50 backbone and 8 data loader workers. The learning rate was set to 0.0026, with decays applied at epochs 43 and 54. Optimization employed momentum (0.9) and weight decay (0.0005). Evaluations were conducted at epochs 21, 32, 37, 42, 48, 53, 59, and 64.

\subsection{FasterRCNN}
Faster R-CNN~\cite{ref_faster_rcnn} was implemented using Detectron2 with a COCO-pretrained ResNet-50-FPN backbone. Training ran for 300 iterations with 8 images per batch and 128 region proposals per image. A constant learning rate of 0.00025 was used without decay, utilizing 2 data loading threads.

\subsection{PPYoloE+}
PP-YOLOE+~\cite{ref_ppyoloe}  employed a CSPResNet backbone and CSPPAN neck, trained for 100 epochs with a batch size of 8 in 32-bit precision. The learning rate schedule included a 5-epoch linear warmup followed by cosine decay from a base of 0.001. SGD was used with momentum 0.9 and L2 regularization of 0.0005. Data augmentations included geometric distortions, cropping, flipping, and multi-scale resizing. Synchronized BatchNorm and EMA (decay 0.9998) were also applied.

\subsection{NanoDETm+}
NanoDetm+~\cite{ref_nanodet} used a ShuffleNetV2 backbone with GhostPAN and LeakyReLU activation. Training was performed over 300 epochs with a batch size of 16 and 4 workers, using AdamW with an initial learning rate of 0.001 and weight decay of 0.05. A 500-step linear warmup from 0.0001 preceded a cosine annealing schedule down to 0.00005. Augmentations included scaling (0.6–1.4×), stretching (0.8–1.2×), translation (±0.2), brightness, contrast, saturation, and horizontal flipping. Gradient clipping (max norm 35) and validation every 10 epochs were applied.


\subsection{Sparse RCNN} 
Sparse R-CNN~\cite{ref_sparse_rcnn} was trained with Detectron using an ImageNet-pretrained ResNet-50 backbone. The model produced 50 region proposals per image and was configured for 8 object classes. Training used the AdamW optimizer with a base learning rate of 0.0001, a batch size of 4, and gradient clipping (norm 1.0). The schedule planned for 220,000 iterations with decays at 150,000 and 200,000, but training was halted at 6,000 iterations due to early convergence. Evaluation occurred every 2,000 iterations.

\subsection{RetinaNET}
The RetinaNet~\cite{ref_retinanet} model was trained using and initialized with COCO-pretrained ResNet-50-FPN backbone without loading pretrained weights. The training used 2 worker threads and processed 8 images per batch. A total of 300 iterations were performed, with a base learning rate of 0.00025 and no scheduled decay steps. Each image had 128 region proposals during training.

\subsection{Yolov12}
YOLOv12x~\cite{ref_yolov12}  was trained for 100 epochs with a batch size 8, and 32-bit precision with AMP. A 3-epoch warmup was used, linearly increasing momentum (0.8→0.937) and bias learning rate (up to 0.1). The base learning rate was 0.01 with no cosine decay, and weight decay was 0.0005. Data augmentation included HSV jittering (h=0.015, s=0.7, v=0.4), horizontal flipping (50\%), translation (±10\%), and scaling (0.5×).

\section{Observations}
Comparing the results from our models with their COCO benchmarks reveals a marked deterioration in performance on our feature-deficient dataset for most of the models. For instance, PP-Yoloe+ exhibits a 15.9 percentage point decline in Average Precision (AP), and YOLOv12 falls from 55.2 mAP to 38.9 mAP. 

\begin{table}[H]
\centering
\caption{Comparison of model performance across different metrics.}
\label{tab:performance}
\begin{tabular}{|c|c|c|c|c|c|c|c|}
\hline
\textbf{Model} & \textbf{Backbone} & \textbf{AP} & \textbf{AP\textsubscript{50}} & \textbf{AP\textsubscript{75}} & \textbf{AP\textsubscript{S}} & \textbf{AP\textsubscript{M}} & \textbf{AP\textsubscript{L}} \\
\hline
SSD300 & ResNet-50 & 29.71 & 82.14 & 13.59 & 6.76 & 31.08 & 8.530 \\
\hline

RetinaNet & ResNet-50-FPN & 1.78 & 6.43 & 0.22 & 0.0 & 2.07 & 2.57 \\
\hline

Faster R-CNN & ResNet-50-FPN & 14.73 & 45.89 & 4.61 & 4.02 & 14.10 & 6.97 \\
\hline

YOLOv12x & R-ELAN & 38.9 & 86.7 & - & - & - & - \\
\hline

PP-YOLOE+x & CSPResNet & 26.8 & 71.7 & 12.0 & 2.2 & 26.4 & 11.7 \\
\hline

NanoDet+m & ShuffleNetV2 & 27.6 & 74.42 & 10.69 & 7.18 & 27.46 & 13.04 \\
\hline

Sparse R-CNN  & ResNet-50 & 25.3 & 72.4 & 10.1 & 10.3 & 26.1 & 12.1 \\
\hline
\end{tabular}
\end{table}

From the benchmarking results, it can be inferred that deeper models with higher complexity, such as Sparse R-CNN and transformer-based architectures, did not consistently outperform comparatively simpler models like YOLO. This suggests that the limitations in detection performance may not be addressed solely by increasing the depth or complexity of the model. Instead, other factors such as data quality, training strategy, or architectural suitability can play a more critical role in improving detection accuracy.

These outcomes underscore the limitations of conventional object detection models when applied to feature-deficient datasets, such as those composed of astrophotography images. The extremely low signal-to-noise ratio, combined with minimal spatial detail for the objects of interest, presents a significant challenge for standard detection architectures.

\begin{figure}[H]
  \centering
  \includegraphics[scale=0.45]{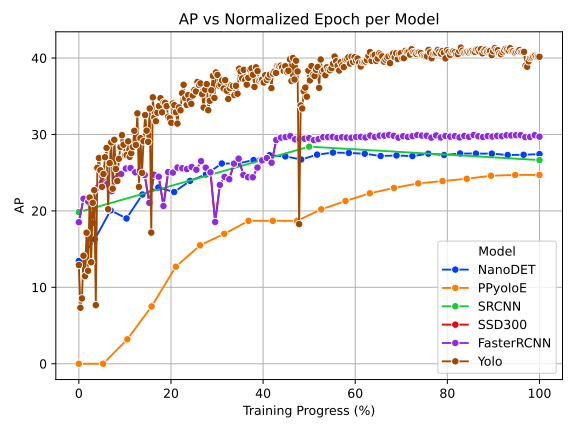}
  \caption{Comparative accuracies over all the models with normalized epochs}
\end{figure}

These findings underscore the need for a specialized detection pipeline optimized for low-feature environments—integrating advanced denoising, robust preprocessing, and region-focused detection tailored to sparsely activated pixels. Additionally, the current limitation of only eight labeled classes contributes to higher misclassification rates, especially for smaller or visually similar celestial bodies (see Fig.~8).

\begin{table}
\centering
\caption{Performance comparison of models on different datasets.}
\label{tab_mAP}
\begin{tabular}{|c|c|c|c|c|c|c|}
\hline
\textbf{Models} & \multicolumn{2}{c|}{\textbf{MobilTelesco}} & \multicolumn{2}{c|}{\textbf{COCO2017}} & \multicolumn{2}{c|}{\textbf{Pascal VOC2012}} \\
  & $\text{mAP}_{50:95}^{\text{val}}$ (\%) & \shortstack{Batch\\size}  & $\text{mAP}_{50:95}^{\text{val}}$ (\%) & \shortstack{Batch\\size}  & $\text{mAP}_{50:95}^{\text{val}}$ (\%) & \shortstack{Batch\\size}  \\
\hline
SSD300 & 29.7 & 32 & 23.2~\cite{ref_ssd} & 32 & 72.4 ~\cite{ref_ssd} & 32 \\
\hline
RetinaNet & 1.78 & 8 & 37.4~\cite{ref_retinanet,ref_resnet} & 8 & 77.28~\cite{ref_miaod} & 2 \\
\hline
Faster R-CNN & 14.73 & 8 & 37.9~\cite{ref_faster_rcnn} & - & 70.8~\cite{ref_improved_faster_rcnn} & - \\
\hline
YOLOv12x & 38.9 & 8 & 55.2~\cite{ref_yolov12} & 32 & - & - \\
\hline
PP-YOLOE+x & 26.8 & 8 & 54.7~\cite{ref_ppyoloe} & 64 & - & - \\
\hline
NanoDet+m & 27.57 & 16 & 30.4~\cite{ref_nanodet} & - & - & - \\
\hline
Sparse R-CNN & 25.3 & 4 & 42.8~\cite{ref_miao2023}& 16 & - & - \\
\hline
\end{tabular}
\end{table}
\vspace*{10pt}  
\noindent
\section{Conclusion}
The results indicate a significant degradation in detection performance when state-of-the-art models trained on general-purpose datasets like COCO are applied to feature-deficient datasets such as MobilTelesco. For example, YOLOv12x shows a drop in mAP from 55.2\% to 38.9\%, and RetinaNet falls to 1.78\% [Refer Table ~\ref{tab_mAP}]. This performance gap suggests that conventional detectors are insufficient under such extreme imaging conditions, motivating several avenues for future research some of which are listed hereby.

\begin{figure}[H]
  \centering
\includegraphics[width=\textwidth]{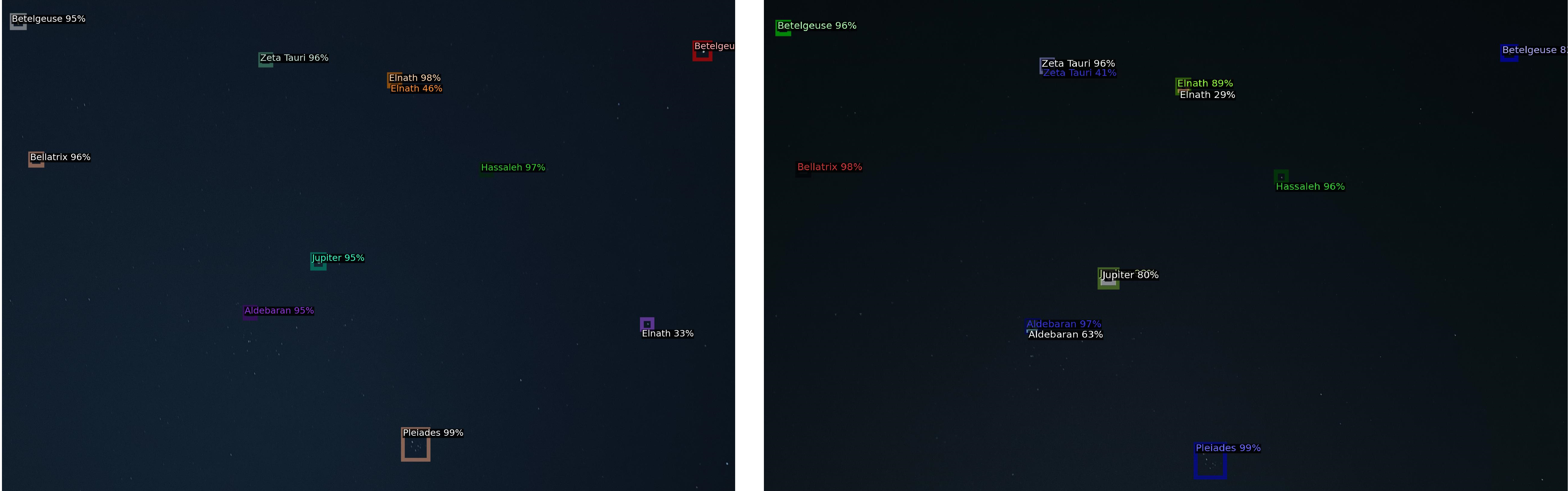} 
  \caption{Some results from inference using Yolo and SSD}
\end{figure}

Advanced Preprocessing and Architectures: To improve detection on low-SNR images,image preprocessing techniques such as BM3D denoising and learned noise suppression networks~\cite{ref_bm3d} can be incorportated. Specifically, we plan to explore multi-scale detection heads and region-focused subnetworks designed to capture sparse, high-contrast objects~\cite{ref_astrometrynet}.

Synthetic Data Formation: Since annotated data in this domain is limited, we plan to pretrain the models on synthetically generated star fields with realistic noise using GANs and astrophysical simulations~\cite{ref_ganastro,ref_transferastro}. 

Segmentation as an Alternative: Instead of bounding-box detection, future work will investigate semantic and instance segmentation approaches which are better suited to distributed or overlapping patterns such as constellations. Dense networks like Tiramisu~\cite{ref_tiramisu} have shown promise under high noise, and weakly supervised segmentation may offer a practical middle ground given the annotation cost of full masks.

Constellation-Based Supervision: An alternative to Synthetic Data is to representing constellations through annotated lines or keypoints, modeling their geometric layout via graph-based networks. This could also allow keypoint regression or edge-detection formulations~\cite{ref_graphconstellation}. 

These directions directly address the bottlenecks exposed by our benchmarks and provide a roadmap for enabling robust astronomical object detection using modest hardware under extreme conditions.


\section*{Acknowledgments}
The author gratefully acknowledges Ms. Maria Pasayalo (University of Florida) for her support in setting up the collaborative LaTeX environment. Special thanks are also extended to Ms. Sophia Bhatti (University of Pennsylvania) for her assistance in manually annotating a portion of the dataset using the LabelImg tool, following the author's labeling guidelines.

\end{document}